\documentclass[9.5pt,journal,compsoc]{IEEEtran}
\usepackage{multirow}
\usepackage{makecell}
\usepackage{tablefootnote}
\usepackage[table]{xcolor}
\usepackage{tikz}

\usepackage{colortbl}

\usepackage{float}
\usepackage{booktabs}
\usepackage{hyperref}
\usepackage{pgfplots}
\usepackage{pgfplotstable}
\pgfplotsset{compat=1.18} % Ensure compatibility

\ifCLASSOPTIONcompsoc
  \usepackage[nocompress]{cite}
\else
  \usepackage{cite}
\fi
\ifCLASSINFOpdf
\usepackage{amsmath}
  \usepackage[caption=false,font=footnotesize,labelfont=sf,textfont=sf]{subfig}
  \usepackage[caption=false,font=footnotesize]{subfig}
\usepackage{fixltx2e}
\usepackage{stfloats}
\begin{document}
\title{Pairwise Discernment of AffectNet Expressions with ArcFace}

\author{Dylan~Waldner,  Shyamal~Mitra      % <-this % stops a space
\IEEEcompsocitemizethanks{\IEEEcompsocthanksitem S. Mitra is with the Department
of Computer Science, The University of Texas at Austin, Austin,
TX, 78705.\protect\\
E-mail: see https://www.cs.utexas.edu/$\sim$mitra/
\IEEEcompsocthanksitem Dylan Waldner was an undergrad at The University of Texas at Austin.}% <-this % stops an unwanted space
\thanks{Manuscript received September 19, 2024; revised Month,  Day, Year.}}


\markboth{Journal of \LaTeX\ Class Files,~Vol.~XX, No.~X, Month~Year}%
{Shell \MakeLowercase{\textit{et al.}}: Bare Demo of IEEEtran.cls for Computer Society Journals}



\IEEEtitleabstractindextext{%
\begin{abstract} This study takes a preliminary step toward teaching computers to recognize human emotions through Facial Emotion Recognition (FER). Transfer learning is applied using ResNeXt, EfficientNet models, and an ArcFace model originally trained on the facial verification task, leveraging the AffectNet database—a collection of human face images annotated with corresponding emotions. The findings highlight the value of congruent domain transfer learning, the challenges posed by imbalanced datasets in learning facial emotion patterns, and the effectiveness of pairwise learning in addressing class imbalances to enhance model performance on the FER task. \end{abstract}

}

\maketitle

\IEEEdisplaynontitleabstractindextext

% \fi
\IEEEpeerreviewmaketitle

\IEEEraisesectionheading{\section{Introduction}\label{sec:introduction}}

\IEEEPARstart{D}{eep} learning for visual classification tasks takes advantage of contemporary explosions in computing power. As computing power has grown over the past 40 years, so have the applications for computer vision tasks, including but not limited to Facial Recognition, Object Detection, Handwriting Authentication, Autonomous Driving, and Image Generation. We will focus on the growing field of Facial Emotion Recognition (FER), which has made significant strides in controlled settings (models posing) but still faces substantial challenges in tasks "in the wild," where lighting, camera angles, and resolution are not optimized. While FER certainly has applications in surveillance and security, this project is the first stepping stone towards work in communicating emotion to machines.

This section will review the history of Image classification and Facial Emotion Recognition. Section~\ref{sec:Data} will describe the strengths and limitations of our dataset and the manipulation tactics we employed. Section~\ref{sec:tools} will explain the results of our experiments. Section~\ref{sec:results} will describe the results of our experiments. Section~\ref{sec:conclusion} will detail sources of error and the short and long-term trajectories of Facial Emotion Recognition.
%I wish you the best of success.
%\IEEEpubidadjcol

\subsection{History}
Deep learning for image classification traces its origins to the Neocognitron \cite{neocognitron}, developed by Kunihiko Fukushima in 1980. The Neocognitron was modeled after Hubel and Wiesel’s studies on human visual processing \cite{HubelWiesel1962}\cite{HubelWiesel1965}\cite{HubelWiesel1977}, which described simple and complex cells that capture patterns across multiple layers of abstraction. It implemented a network that processed small patches of input images and passed data from simple to complex cells repeatedly to form S-layers, or "feature maps," which represent specific patterns or features and ultimately produced an image classification output.

While the Neocognitron is widely considered the first Convolutional Neural Network (CNN), Yann LeCun’s LeNet \cite{LeNet} expanded on its design to establish the modern CNN architecture. LeNet processed 32x32 input images of hand-drawn numbers [0-9] and classified them as the correct integer with 99\% accuracy. LeNet accomplished this by introducing a sliding filter with shared weights that "convolved" over the input data, enhancing pattern recognition robustness through invariant feature detection. This sliding layer has become the foundation for the modern CNN. Building on the Neocognitron’s data abstraction, LeNet incorporated pooling layers, where subsampling captured more general patterns in the image and reduced overfitting. LeNet’s most lasting contribution was the addition of a fully connected layer—a dense matrix operation where input weights are connected to all output weights. LeCun included the fully connected layer in anticipation of advances in computing power, which he believed would enable dense layers to detect intricate patterns. However, the limitation at the time was that contemporary computing power remained relatively constrained, and with it, so did LeNet’s broader capabilities.

As computing power increased, GPU performance and availability breakthroughs led to an explosion of deep learning models for image classification in the mid-2010s. At the forefront was AlexNet \cite{AlexNet}, which processed 224x224 input images from ImageNet and achieved a top-5 classification accuracy rate of 85\%. AlexNet introduced several innovations that became foundational in contemporary deep learning, including the ReLU activation layer, which enabled non-linear decision boundaries; Local Response Normalization, a precursor to Batch Normalization that enhanced contrast in feature maps; Dropout, a regularization technique that randomly deactivated neurons during training to reduce overfitting; and data augmentation techniques combined with multi-GPU architectures that set new standards for scalability and performance.

However, AlexNet falls short compared to modern image classification models due to the computing limitations of its time, which restricted the network to 8 layers and constrained its ability to identify complex patterns. Despite these limitations, AlexNet paved the way for a series of models that competed for top performance on the ImageNet data set \cite{imagenet}, processing 224x224 input images and classifying them into 1 of 5,247 categories. Network in Network \cite{NetworkinNetwork} introduced non-linearity within the filters by incorporating Multi-Layer Perceptrons (MLPs), enhancing the network’s expressiveness. VGGNets \cite{VGGNets} revamped AlexNet’s hyper-parameters and restructured the GPU architecture into an ensemble, doubling the network’s depth while maintaining the same number of parameters. GoogLeNet \cite{GoogleNet}, named in homage to LeNet, introduced the inception block, which improved computational efficiency and pattern recognition by leveraging dimension reduction and sparse matrices, enabling neurons to focus on capturing the most relevant features. GoogLeNet also introduced softmax classification blocks at intermediate stages of the network architecture to reinforce gradient strength and address the vanishing gradient problem.

ResNet \cite{he2015deep} revolutionized deep learning by leveraging identity mapping, which adds the input of a block directly to its output. This technique allowed the model to focus on the residual—or the change—within an image as it passes through the model, leading to more focused and manageable learning. A key benefit of residual mapping was improved gradient strength, which, combined with GoogLeNet’s intermediary classification blocks, effectively mitigated the vanishing gradient issue. The overarching theme of these mid-decade models was achieving greater network depth while maintaining a fixed number of parameters, resulting in higher accuracies. However, despite their strong performance in classification tasks, these models were so complex and memory-intensive that deployment costs and test speeds remained prohibitive for widespread use.

The next wave of CNNs for image classification prioritized optimization to address the high memory and computational costs of earlier models. Like their predecessors, these models were trained on ImageNet, processing 224x224 input images, but they achieved comparable accuracy with significantly reduced memory usage and computational demands. Notable examples include SqueezeNet \cite{squeezenet}, Squeeze-and-Excite \cite{squeezeandexcitation}, MobileNet \cite{mobilenetv1} \cite{mobilenetv2}, and EfficientNet \cite{efficientnet}. These models focused on compressing model size and complexity to make CNNs more practical for real-world applications by optimizing the relationship between model architecture and the hardware they operate on.

SqueezeNet introduced techniques such as 8-bit quantization on sparse matrices, optimizing the sparse matrices introduced by GoogLeNet for GPUs, which specialize in dense matrix multiplication. MobileNet leveraged the relationship between parameter count and kernel dimension by splitting standard convolutional layers into \textit{Depthwise} and \textit{Pointwise} filters. This innovation maintained accuracy while cutting memory usage by a factor of \((\frac{1}{N} + \frac{1}{D_K^2})\), where \(N\) is the number of output channels and $D_K$ is the kernel dimension. These advances marked a shift toward creating lightweight, efficient architectures suitable for deployment on devices with limited computational resources, such as smartphones and embedded systems.

In the modern era, the focus of image classification has shifted from relying solely on CNNs to incorporating Vision Transformers (ViTs) and hybrid models. This transition was pioneered by the seminal paper "Attention Is All You Need" \cite{attention}, which introduced transformers as a framework for handling sequential data. Building on this foundation, Google repurposed transformers for image classification in their groundbreaking work \cite{ViT}, driven by the intuition that the transformer’s superior computational efficiency and scalability could allow it to perform equivalently to, or even surpass, CNNs in visual recognition tasks.

\subsection{Research Statement}
The goal of our research was to differentiate between emotions in images of peoples’ faces. We used the AffectNet dataset and three different Convolutional Neural Networks. We employed transfer learning on the CNNs to classify images into one of the eight AffectNet classes. Our study has implications for research in facial emotion recognition, human-computer interaction, cognitive behavioral therapy, image and video generation, and security. 

\section{Data} \label{sec:Data}

We used the AffectNet Database \cite{affectnet} to train our models and run experiments. More specifically, we selected a subset of the AffectNet database consisting of \(\approx\) 281,000 images, each annotated in four different ways: Expression, Arousal, Valence, and Landmarks. Expression assigned a label corresponding to the AffectNet emotion classes: [Neutral, Happy, Sad, Disgust, Fear, Anger, Surprise, and Contempt]. Arousal assigned a value in \([-1, +1]\), reflecting whether an expression is exciting/agitating or calm/soothing. Valence assigned a value in \([-1 +1]\), reflecting how positive or negative an expression is (e.g., happy vs. sad, angry vs. neutral). Landmark annotated the features on each image’s face, providing a numerical representation of facial features. The larger AffectNet dataset included classes "Non-Face," "None," and "Uncertain." We cleaned our dataset of these classes, yielding a much more refined dataset on which to perform classification.

We chose the AffectNet database for its size, its annotations, and the diversity of people represented. The AffectNet database was constructed from images scraped from Google Images, meaning they were photos posted by ordinary people online, rather than professionally taken images. This type of photo is referred to as "in the wild." This placed work using the database at the forefront of the FER challenge due to the inherent noise in the dataset caused by variations in picture angle and lighting. The challenge was further compounded by the imbalance of emotional representation "in the wild"; people were more likely to post happy photos of themselves and others online than photos depicting sadness or disgust. As a result of this trend, the cleaned AffectNet dataset we worked with had the class distribution summarized in Table~\ref{tab:expression_classes}.

\begin{table}[h!]
    \centering
    \caption{AffectNet Expression Classes, Counts, and Proportions}
    \begin{tabular}{lrr}
        \toprule
        \textbf{Expression Class} & \textbf{Count} & \textbf{Proportion (\%)} \\ 
        \midrule
        Neutral   & 74874  & 26.02 \\
        Happy     & 134415 & 46.72 \\
        Sad       & 25459  & 8.85  \\
        Surprise  & 14090  & 4.90  \\
        Fear      & 6378   & 2.22  \\
        Disgust   & 3803   & 1.32  \\
        Anger     & 24882  & 8.65  \\
        Contempt  & 3750   & 1.30  \\
        \bottomrule
    \end{tabular}
    \label{tab:expression_classes}
\end{table}

Dealing with this significant skew was challenging because the dataset accurately represented the distribution of images on the internet. A generalized model deployed on the images "in the wild" needed to account for this imbalance. However, this class imbalance disrupts the model from learning the underlying facial emotion feature patterns, preventing true Facial Emotion Recognition. In Section~\ref{sec:results} we discuss our solution to this contradiction.

A significant limitation of the dataset was training a model to sufficiently differentiate between minority-represented classes and between a minority represented class and a majority class. For example, differentiating between fear, disgust, and contempt posed a major challenge, as these three classes combined to just 14,000 photos, representing only \(\approx\) 5\% of the dataset. This meant that a model trained on this dataset had a bias towards \(\approx\) 95\% of its data that from the five major classes and needed to learn to differentiate between three similar classes that comprised only \(\approx\) 5\% of the dataset. In practice, the model tended to over-predict the majority classes, often classifying minority classes as one of the majority classes. In this paper, we aimed to generalize the model to the underlying patterns associated with each emotion independent of the noise and imbalance that results from "in the wild" data.

\subsection{ResNeXt and EfficientNet}
The ResNeXt101-32x4d \cite{resnext} and EfficientNet-b0 \cite{efficientnet} models we used were originally trained on the ImageNet dataset \cite{imagenet}. ImageNet consisted of 3.2 million images from the internet, spanning 5,247 classes. This was achieved through a massive collection of internet images, which humans then verified using Amazon Mechanical Turk. Multiple humans verified each image to reach a consensus, particularly for niche classes. We applied transfer learning to these models, transitioning from a general image classification task to the facial emotion recognition task.

\subsection{ArcFace}
The ArcFace model we used was trained on Microsoft’s MS1MV2 facial verification dataset \cite{ms1m}. Composed of \(\approx\) 10 million images of \(\approx\) 100,000 celebrities,  the dataset included individuals from 2,000 different professions across 200 countries, with a wide range of ages and a higher proportion of women compared to men\footnote{The paper dismisses the imbalance claiming the genders can be balanced manually by keyword search.}. Images were scraped from the web using keyword searches and were labeled by the MS1M authors. We applied transfer learning to this model, facilitating a more natural transition from the facial verification task to the facial emotion recognition task.

\subsection{Related Works}

Our work followed the trend of Facial Recognition tasks in Image Classification. We built on previous experiments that utilized the AffectNet database, including studies by Ngo and Yoon \cite{weightedcluster}, Deng et al. \cite{deng2019arcface}, and Siddiqui et al. \cite{siddiqui2022robust}. 

Ngo and Yoon introduced the weighted-cluster loss function, which aimed to improve upon weighted loss functions designed for imbalanced datasets. Among these weighted loss functions was weighted softmax, which penalized misclassifications of minority classes more than majority classes. However, weighted softmax fell short because it could not handle high inter-class similarities and intra-class variations. Ngo and Yoon demonstrated that center loss could be incorporated alongside softmax to reduce intra-class variation by drawing samples closer to their class center. Center loss also fell short, however, as major classes were updated more frequently than minor classes, limiting its ability to tighten minority clusters effectively.

Ngo and Yoon showed that weighted-cluster loss addressed the imbalanced class problem left by center loss by adding a new term that simultaneously pulled the centers of each cluster apart while drawing each sample within the cluster closer to its center. Their paper demonstrated that weighted-cluster loss made each cluster more compact and distinct, which should have resulted in improved classification. They deployed a weighted-softmax function on the rearranged cluster feature space to classify the logits into a final prediction.

Building on Ngo and Yoon’s weighted-cluster loss function, we extended our research to ArcFace and Angular Additive Margin \cite{deng2019arcface}. Angular Additive Margin followed the precedent set by Triplet loss and Center loss, enhancing intra-class compactness and increasing inter-class separation in the feature space. Deng et al. demonstrated that Angular Additive Margin rearranged the feature space before classification, simplifying the drawing of decision boundaries by creating more distinct and well-defined features.

Angular Additive Margin (AAM) is derived from the softmax function:

\begin{equation}
L_{SoftMax} = - \log \frac{e^{W_{y_i}^T x_i + b_{y_i}}}{\sum_{j=1}^{N} e^{W_j^T x_i + b_j}}
\end{equation}

and defined by:

\begin{equation}
    L_{AAM} = - \log \frac{e^{s \cos(\theta_{y_i} + m)}}{e^{s \cos(\theta_{y_i} + m)} + \sum_{j=1, j \neq y_i}^{N} e^{s \cos \theta_j}}
\end{equation}

In \textbf{Softmax}, the term \( W_j^T x_i \) measures the similarity between the weight vector \( W_j \) and the feature vector \( x_i \). In \textbf{ArcFace}, this term is modified to \( ||W_j|| ||x_i|| \cos(\theta_j) \), where both \( W_j \) and \( x_i \) are normalized to 1, simplifying to \( \cos(\theta_j) \). ArcFace introduced two additional parameters: \( m \), an additive angular margin that increased the separation between classes in the embedding space, and \( s \), the scaling factor that controlled the radius of the hypersphere to which the features are normalized and projected onto.  Normalizing the features to the surface of the hypersphere allowed predictions to depend only on the angle between the weights \(W\) and \(x\) the features, measured by \( \cos(\theta_j) \). When this feature space was passed on to the final classification, the feature vectors were more distinct from each other, making decision boundaries easier to draw. 

Finally, we followed the baselines set by the AffectNet paper \cite{affectnet}, which we will discuss in Section~\ref{sec:tools}.

\section{Tools For Analysis} \label{sec:tools}

We aimed to improve the baselines established by the AffectNet paper by implementing and combining various approaches into a single model. Initially, we conducted a series of experiments with a ResNeXt101-32x4d CNN \cite{resnext} and then transitioned to an EfficientNet-B0 CNN\cite{efficientnet} from TorchHub, which demonstrated higher efficiency and accuracy compared to ResNeXt. We removed the final fully connected layers from both models and replaced them with four layers in a multi-output module, each employing a fully connected layer for a specific annotation: arousal, valence, expression, and landmarks. We froze the weights for the entire model except for the expression classifier. The losses from the four output layers were added together to train a fully connected classification layer for expression. We found that model accuracy decreased when only the expression classifier’s loss was used. In our initial experiments, we tested three different loss functions as outlined in the original AffectNet paper: Mean Squared Error (MSE), Pearson Correlation Loss, and Signed MSE Loss. \cite{affectnet}

Both models were pre-trained on ImageNet, so we normalized the images based on ImageNet’s standards of mean and standard deviation (std): mean = [0.485, 0.456, 0.406] and std = [0.229, 0.224, 0.225]. To match the expected input dimensions, we resized the images to 224x224 for the ResNeXt and EfficientNet models. We applied random horizontal flips as well as random horizontal and vertical translations to augment the data for minority classes. We used the ADAM optimizer with a Reduce On Plateau (ROP) learning rate scheduler. For ResNeXt and EfficientNet, the learning rate scheduler was configured to start at 0.128 and decrease by a factor of 10 when the training loss plateaued for five epochs.

After the initial experiments, we switched to a ResNet100 model pre-trained on the facial recognition dataset MS1MV2. This model was trained using Angular Additive Margin, which was specifically designed for the facial recognition task—a task inherently more complex than facial emotion recognition. For this model, we initially set the learning rate scheduler to the same parameters as the ResNeXt and EfficientNet models. However, through experimentation, we adjusted to an initial learning rate of 0.01 with a patience of 5 and a factor of 0.25. We once again experimented with data augmentation techniques, including color jitter, random transformations, rotations, and reflections. Input images were scaled to 112x112 to match the expected input dimensions.  

We experimented with calculating the valence, landmark, and arousal loss using Signed Mean Squared Error and Pearson Correlation Loss. Signed Mean Squared Error accounted for the direction of the error, which was critical, particularly for valence and arousal, as their direction determined the quality of the expression (e.g., excitement, boredom) and the magnitude of the assigned value represented the intensity of the emotion. For this reason, for an actual value of 0.3, a predicted value of -0.1 was not as accurate as a predicted value of 0.7, even though the absolute distance between the points was identical. Pearson Correlation Loss took a similar approach by measuring the correlation between the predicted and ground truth values. To function as a loss metric, we subtracted the Pearson Correlation from 1 to maximize the correlation. We determined Signed MSE yielded slightly better results, and so all of our reported results used Signed MSE error.

The ResNeXt model was accessed from TorchHub; the EfficientNet model was accessed from Torchvision. The ArcFace backbone and weights were downloaded from the \href{https://github.com/deepinsight/insightface/blob/master/recognition/arcface_torch/README.md}{InsightFace GitHub Repo}. All training and test runs were done on the UT Servers using HTCondor. 

\section{Results} \label{sec:results}

\subsection{Initial Results}
The initial results of the training loops run on the AffectNet dataset were very promising. Starting at \(\approx\) 50\% validation accuracy for the ResNeXt model, we got up to \(\approx\) 60\% validation accuracy with the EfficientNet model. We moved on from these ImageNet-trained, stock CNNs, and repurposed the pretrained ArcFace model available from InsightFace to leverage its success in Facial Recognition into the Facial Emotion Recognition task. Initial tests were run with 40 epoch training loops and a 0.01 LR that was reduced 10x after five epochs of plateau.

However, what became apparent very early on was the issue the class imbalance in the dataset posed to the model’s robustness. While the largest class, Happiness, boasted a recall \(>\) 92\%, the three least represented classes (Fear, Disgust, and Contempt) regularly had precision, recall, and F1-scores \(<\) 12\%. Table~\ref{tab:arcface_metrics} illustrates these disparities that manifested during training and validation time.

\begin{table}[h!]
    \centering
    \caption{ArcFace Training L2=0.0005 Metrics}
    \label{tab:arcface_metrics}
    
    % Training vs Validation
    \begin{tabular}{@{}lcc@{}}
    \toprule
    \textbf{Metric} & \textbf{Training} & \textbf{Validation} \\ \midrule
    Loss            & 21685.281         & 21655.095           \\
    Accuracy (\%)   & 79.130            & 74.226              \\ \bottomrule
    \end{tabular}
    
    \bigskip % Vertical space between sub-tables
    
    % Overall Metrics
    \begin{tabular}{@{}lccc@{}}
    \toprule
    \textbf{Overall Metric} & \textbf{Precision} & \textbf{Recall} & \textbf{F1 Score} \\ \midrule
    Values                  & 0.724              & 0.742           & 0.724             \\ \bottomrule
    \end{tabular}
    
    \bigskip % Vertical space between sub-tables
    
    % Class-Level Metrics
    \begin{tabular}{@{}lccc@{}}
    \toprule
    \textbf{Class} & \textbf{Precision} & \textbf{Recall} & \textbf{F1 Score} \\ \midrule
    Neutral        & 0.631              & 0.788           & 0.701             \\
    Happy          & 0.869              & 0.925           & 0.896             \\
    Sad            & 0.680              & 0.420           & 0.519             \\
    Surprise       & 0.486              & 0.452           & 0.468             \\
    Fear           & 0.363              & 0.115           & 0.175             \\
    Disgust        & 0.248              & 0.100           & 0.142             \\
    Anger          & 0.660              & 0.481           & 0.556             \\
    Contempt       & 0.074              & 0.013           & 0.022             \\ \bottomrule
    \end{tabular}
\end{table}

To balance the data, we experimented with a weighted random sampler in the dataloader to even out class representations by using the reciprocal of the class count for weights: \(1/size(class)\). We also applied transformations to the minority classes to maintain data diversity, utilizing the Torchvision transforms library to implement these changes. We tested affine transformations with random 10\% horizontal and vertical translations and random 2-degree rotations. Color jitters were configured with a 10\% random brightness change, 20\% random contrast change, 15\% random saturation change, and 5\% random hue change. Random horizontal flips occurred with a 50\% probability. However, despite these efforts, the imbalanced dataset persisted in the model’s representations of facial emotions.

\subsection{Capturing Facial Emotion Patterns}

The various approaches we employed were unsuccessful in enabling the model to accurately capture the underlying facial emotion patterns, primarily due to significant class imbalances in the training data. This limitation was evident in Table~\ref{tab:arcface_metrics}, where the recall scores for majority classes were disproportionately high, highlighting the model’s tendency to overpredict these classes. While the model achieved high accuracies for majority classes, this performance was driven by a strong bias toward those classes rather than a genuine understanding of the facial features associated with emotions like happiness or neutrality. This bias became particularly apparent during testing, where we evaluated the model using the balanced working test set\footnote{The official test set is withheld for competitions} from the AffectNet dataset, as summarized in Table~\ref{tab:expression_classes_combined}.

\begin{table}[t]
    \caption{AffectNet Expression Classes and Their Counts for Skewed and Balanced Test Datasets.}
    \label{tab:expression_classes_combined}
    \centering
    \renewcommand{\arraystretch}{1.3} % Adjust row height for better readability
    \setlength{\tabcolsep}{8pt}      % Adjust column spacing
    \begin{tabular}{l r r}
        \toprule
        \textbf{Expression Class} & \textbf{Balanced Count} & \textbf{Skewed Count } \\ \midrule
        Neutral                   & 500                   & 278                    \\
        Happy                     & 500                   & 500                    \\
        Sad                       & 500                   & 94                     \\
        Surprise                  & 500                   & 52                     \\
        Fear                      & 500                   & 23                     \\
        Disgust                   & 500                   & 14                     \\
        Anger                     & 500                   & 92                     \\
        Contempt                  & 499                   & 13                     \\ \bottomrule
    \end{tabular}
\end{table}

The training dataset exhibited a significant class imbalance, reflecting the biases inherent in how people share images of themselves and others online. However, the test dataset did not account for this imbalance. By equalizing all classes, the test dataset shifted its focus away from evaluating the model’s ability to discern between facial expressions "in the wild" and instead assessed its ability to distinguish facial emotions in isolation. As a result, a significant drop in performance was observed when comparing the training and validation accuracies, where the model learned the "in the wild" task, to the test accuracies, where the model was evaluated on true emotion discernment. These results are summarized in Table~\ref{tab:accuracy_comparison2}.

\begin{table}[t]
    \caption{Accuracy of ArcFace Model on Balanced and Skewed Test Datasets.}
    \label{tab:accuracy_comparison2}
    \centering
    \begin{tabular}{lcc}
        \toprule
        \textbf{Variant} & \makecell{\textbf{Balanced} \\ \textbf{Test Data Set} \\ \textbf{Accuracy (\%)}} & \makecell{\textbf{Skewed} \\ \textbf{Test Data Set} \\ \textbf{Accuracy (\%)}} \\ \midrule
        \makecell{Trained on \\ Artificially \\ Balanced Dataset} & 44 & 67.4 \\ \\
        \makecell{Trained on \\ Naturally \\ Skewed Dataset} & 40.8 & 74.7 \\ \bottomrule
    \end{tabular}
\end{table}

Table~\ref{tab:accuracy_comparison2} presents two distinct accuracy metrics. The balanced test dataset accuracy represents the model’s performance on a dataset with equal representation of all classes, designed to assess its ability to distinguish emotions under standardized conditions. In contrast, the skewed test dataset accuracy evaluates the model’s performance on a dataset with the class distribution shown in Table~\ref{tab:expression_classes_combined}, which was adjusted to reflect the natural class imbalance found in the training data and "in the wild." This skewed test set was specifically designed to measure the model’s effectiveness on the "in the wild" FER task, rather than on purely balanced FER discernment.

To conduct this analysis, we trained and optimized two models. One model was trained and tuned on an artificially balanced AffectNet dataset using a weighted random sampler to equalize class sampling probabilities. A maximum of \(2 \times size(smallestclass)\) samples were selected with replacement. To enhance the diversity of minority classes (Fear, Disgust, and Contempt) and mitigate issues with duplicate samples, these classes were augmented with random affine transformations, including up to 10\% horizontal and vertical translations and random rotations. Additionally, a 50\% probability of horizontal flipping was applied. These augmentations were carefully designed to introduce meaningful diversity without distorting the features of the original images. Our approach maximized the size of the dataset while retaining balanced class representation. 

The model performed significantly better on the "in the wild" test set, confirming its ability to generalize to real-world emotions. However, its performance on the balanced test dataset was similarly poor regardless of whether it was trained on the natural dataset or an artificially balanced dataset. This lack of improvement indicates that broadly re-balancing the training set is insufficient for training the model effectively on the FER task. Re-balancing efforts failed because reducing the training data to equally represent minority classes resulted in an overly condensed dataset, leaving insufficient data for the model to generalize effectively to facial emotion features. Addressing these discrepancies between different facial emotion features required a more targeted and radical approach.

\subsection{Pairwise Discernment}

Due to the massive class imbalances, we determined that a model trained on the AffectNet dataset was best suited to handle pure emotion discernment by distinguishing between pairs of emotions at a time. The motivation for pairwise discernment is inspired by the way people can more accurately identify or describe a musical note when comparing it directly to another, highlighting the power of relative distinctions over isolated recognition. We grouped emotions into pairs (e.g., happy and neutral, happy and disgusted) and evaluated the model trained with the ArcFace architecture fine-tuned on the AffectNet dataset. To balance the classes within each pair, we weighed them equally and performed random sampling \(2 \times min(class1size,class2size)\) times, ensuring equal representation between classes in each pair. While the results still reflected the biases of the generally trained model toward the larger classes, pairwise discernment proved to be a more effective task for the model. The results are summarized in Table~\ref{tab:paircomp}.

\begin{table}[h]
\centering
\caption{Difference in Accuracy Between Single Fully Connected Layer Pair Discernment Model and Fully Connected Pair Dictionary Test Results}
\label{tab:paircomp}
\begin{tabular}{lccl}
\toprule
\makecell{\textbf{Class} \\ \textbf{Pair}} & \makecell{\textbf{One FC Layer} \\ \textbf{Accuracy (\%)}} & \makecell{\textbf{FC Pair Dict} \\ \textbf{Accuracy (\%)}} & \makecell{\textbf{Difference} \\ \textbf{(\%)}} \\
\midrule
Fear      + Contempt         & 92.5                                   & 70.6                                   & -21.9                           \\
Happy     + Sad              & 91.1                                   & 92.3                                   & +1.2                            \\
Happy     + Anger            & 90.5                                   & 92.1                                   & +1.6                            \\
Neutral   + Happy            & 90.4                                   & 90.3                                   & -0.1                            \\
Surprise  + Anger            & 87.1                                   & 73.8                                   & -13.3                           \\
Fear      + Disgust          & 87.1                                   & 65.4                                   & -21.7                           \\
Sad       + Surprise         & 86.9                                   & 77.7                                   & -9.2                            \\
Disgust   + Contempt         & 86.8                                   & 59.5                                   & -27.3                           \\
Happy     + Surprise         & 85.1                                   & 88.8                                   & +3.7                            \\
Sad       + Anger            & 81.5                                   & 67.6                                   & -13.9                           \\
Happy     + Fear             & 81.2                                   & 80.7                                   & -0.5                            \\
Surprise  + Disgust          & 76.4                                   & 52.0                                   & -24.4                           \\
Neutral   + Surprise         & 75.5                                   & 80.2                                   & +4.7                            \\
Fear      + Anger            & 75.5                                   & 59.8                                   & -15.7                           \\
Neutral   + Sad              & 75.3                                   & 76.6                                   & +1.3                            \\
Surprise  + Contempt         & 75.3                                   & 52.3                                   & -23.0                           \\
Neutral   + Anger            & 75.1                                   & 75.6                                   & +0.5                            \\
Happy     + Disgust          & 74.1                                   & 71.3                                   & -2.8                            \\
Sad       + Contempt         & 73.9                                   & 50.9                                   & -23.0                           \\
Anger     + Contempt         & 72.9                                   & 52.7                                   & -20.2                           \\
Neutral   + Fear             & 69.1                                   & 67.1                                   & -2.0                            \\
Neutral   + Disgust          & 67.9                                   & 63.2                                   & -4.7                            \\
Sad       + Fear             & 67.1                                   & 55.5                                   & -11.6                           \\
Sad       + Disgust          & 66.6                                   & 50.9                                   & -15.7                           \\
Surprise  + Fear             & 61.1                                   & 50.5                                   & -10.6                           \\
Disgust   + Anger            & 59.5                                   & 52.3                                   & -7.2                            \\
Neutral   + Contempt         & 54.1                                   & 53.4                                   & -0.7                            \\
Happy     + Contempt         & 52.7                                   & 59.2                                   & +6.5                            \\
\bottomrule
\end{tabular}
\end{table}

To further optimize pairwise discernment, we implemented a dictionary of fully connected (FC) layers, each assigned to a specific pair, to learn the nuances of the differences between pairs. The first implementation utilized the generally trained model by appending the pair-specific FC layer to the general FC layer. The second implementation removed the FC layer fine-tuned on AffectNet, operating under the intuition that the general model’s learned biases could influence the pairwise FC layers. Both implementations were trained for 30 epochs with a learning rate of 0.0001, a learning rate scheduler that reduced the rate by a factor of 0.25 after a five-epoch plateau, a weight decay of 0.0005, a batch size of 256, and the ADAM optimizer.

Theoretically, using a dictionary of FC layers—one dedicated to each class pair—should result in a better architecture for detecting true facial emotion features. This design focuses exclusively on the two classes being compared, enhancing the model's ability to distinguish between them. In contrast, the single FC layer model performs a standard forward pass, extracting logits for all classes but only considers the logits for the relevant pair during evaluation. 

\begin{table}[H]
\centering
\caption{Class-Specific Statistics for Each Pair ArcFace Model}
\label{tab:class_specific}
\begin{tabular}{cccccc}
\toprule
\textbf{Class} & \textbf{Accuracy (\%)} & \textbf{Precision} & \textbf{Recall} & \textbf{F1 Score} \\
\midrule
Fear      & 91.8 & 0.9310 & 0.9180 & 0.9245 \\
Contempt  & 93.2 & 0.9190 & 0.9319 & 0.9254 \\
\midrule
Happy     & 99.6 & 0.8513 & 0.9960 & 0.9180 \\
Sad       & 82.6 & 0.9952 & 0.8260 & 0.9027 \\
\midrule
Happy     & 99.6 & 0.8426 & 0.9960 & 0.9129 \\
Anger     & 81.4 & 0.9951 & 0.8140 & 0.8955 \\
\midrule
Neutral   & 85.4 & 0.9489 & 0.8540 & 0.8989 \\
Happy     & 95.4 & 0.8673 & 0.9540 & 0.9086 \\
\midrule
Surprise  & 88.0 & 0.8644 & 0.8800 & 0.8722 \\
Anger     & 86.2 & 0.8778 & 0.8620 & 0.8698 \\
\midrule
Fear      & 86.2 & 0.8778 & 0.8620 & 0.8698 \\
Disgust   & 88.0 & 0.8644 & 0.8800 & 0.8722 \\
\midrule
Sad       & 86.8 & 0.8697 & 0.8680 & 0.8689 \\
Surprise  & 87.0 & 0.8683 & 0.8700 & 0.8691 \\
\midrule
Disgust   & 85.8 & 0.8755 & 0.8580 & 0.8667 \\
Contempt  & 87.8 & 0.8605 & 0.8778 & 0.8690 \\
\midrule
Happy     & 98.8 & 0.7755 & 0.9880 & 0.8690 \\
Surprise  & 71.4 & 0.9835 & 0.7140 & 0.8273 \\
\midrule
Sad       & 79.4 & 0.8288 & 0.7940 & 0.8110 \\
Anger     & 83.6 & 0.8023 & 0.8360 & 0.8188 \\
\midrule
Happy     & 100.0 & 0.7267 & 1.0000 & 0.8418 \\
Fear      & 62.4 & 1.0000 & 0.6240 & 0.7685 \\
\midrule
Surprise  & 76.4 & 0.6891 & 0.9620 & 0.8030 \\
Disgust   & 56.6 & 0.9371 & 0.5660 & 0.7057 \\
\midrule
Neutral   & 75.5 & 0.6803 & 0.9620 & 0.7970 \\
Surprise  & 54.8 & 0.9352 & 0.5480 & 0.6910 \\
\midrule
Fear      & 75.5 & 0.9443 & 0.5420 & 0.6887 \\
Anger     & 96.8 & 0.6788 & 0.9680 & 0.7980 \\
\midrule
Neutral   & 96.4 & 0.6779 & 0.9640 & 0.7960 \\
Sad       & 54.2 & 0.9377 & 0.5420 & 0.6869 \\
\midrule
Surprise  & 75.3 & 0.6759 & 0.9720 & 0.7974 \\
Contempt  & 53.3 & 0.9500 & 0.5331 & 0.6829 \\
\midrule
Neutral   & 75.1 & 0.6765 & 0.9620 & 0.7944 \\
Anger     & 54.0 & 0.9343 & 0.5400 & 0.6844 \\
\midrule
Happy     & 100.0 & 0.6588 & 1.0000 & 0.7943 \\
Disgust   & 48.2 & 1.0000 & 0.4820 & 0.6505 \\
\midrule
Sad       & 96.4 & 0.6648 & 0.9640 & 0.7869 \\
Contempt  & 51.3 & 0.9343 & 0.5130 & 0.6624 \\
\midrule
Anger     & 72.9 & 0.5543 & 0.9700 & 0.7055 \\
Contempt  & 49.7 & 0.9254 & 0.4970 & 0.6467 \\
\midrule
Neutral   & 69.1 & 0.6195 & 0.9900 & 0.7621 \\
Fear      & 39.2 & 0.9751 & 0.3920 & 0.5592 \\
\midrule
Neutral   & 67.9 & 0.6095 & 0.9960 & 0.7563 \\
Disgust   & 36.2 & 0.9891 & 0.3620 & 0.5300 \\
\midrule
Sad       & 67.1 & 0.6084 & 0.9600 & 0.7448 \\
Fear      & 38.2 & 0.9052 & 0.3820 & 0.5373 \\
\midrule
Sad       & 66.6 & 0.6043 & 0.9620 & 0.7423 \\
Disgust   & 37.0 & 0.9069 & 0.3700 & 0.5256 \\
\midrule
Surprise  & 61.1 & 0.5668 & 0.9420 & 0.7077 \\
Fear      & 28.0 & 0.8284 & 0.2800 & 0.4185 \\
\midrule
Disgust   & 22.0 & 0.8800 & 0.2200 & 0.3520 \\
Anger     & 97.0 & 0.5543 & 0.9700 & 0.7055 \\
\midrule
Neutral   & 54.1 & 0.5216 & 0.9900 & 0.6832 \\
Contempt  & 9.0 & 0.9000 & 0.0902 & 0.1639 \\
\midrule
Happy     & 52.7 & 0.5139 & 1.0000 & 0.6789 \\
Contempt  & 5.2 & 1.0000 & 0.0521 & 0.0990 \\
\bottomrule
\end{tabular}
\end{table}

However, this approach does not guarantee that the model is learning to differentiate based on meaningful facial emotion features; it could simply produce higher logits for the correct class by chance, even if both logits are low.

In practice, the fully connected layer dictionary approach performed worse than the general model. Table~\ref{tab:paircomp} summarizes the comparison, showing that while highly represented classes benefited from the FC dictionary approach, the pairwise dictionary struggled to effectively learn minority-represented classes. This limitation arose because the dense layers in the pairwise dictionary lacked sufficient data for training on minority class pairs, whereas the general model was trained on the entire AffectNet dataset. As a result, performance dropped by up to 27\%, underscoring the challenges of this approach in addressing class imbalances. 

Table~\ref{tab:class_specific} summarizes the general model's class accuracies for each pair. Since the classes are balanced against each other, high metrics for both classes suggest that the model effectively recognizes genuine facial emotions, reducing the influence of "in the wild" biases. However, not all pairs performed equally well. The pairwise discernment scale provides insight into which emotion pairs the model can effectively differentiate.

Notably, the model consistently distinguished between Happy + Sad faces as well as Happy + Angry faces, achieving strong performance even on new data. An unexpected result was the model's ability to discern between the three minority classes: Fear + Contempt, Fear + Disgust, and Contempt + Disgust. The high metrics for these pairs suggest that the model can reliably differentiate between the minority classes despite the limited data.

\section{Conclusion} \label{sec:conclusion}
Our work introduced a novel approach by applying pairwise discernment to the "in the wild" FER AffectNet dataset, achieving accuracies as high as 92\%. This approach highlighted the challenges of distinguishing multiple emotions simultaneously in images sourced from the internet, where natural biases influence the types of photos people post of themselves and others. Pairwise discernment enabled the model to effectively differentiate between three minority classes (Fear, Contempt, and Disgust) that were indistinguishable during general classification tasks. This enhanced ability to distinguish minority classes suggests that pairwise discernment allows the model to learn underlying facial emotion features that remain elusive in broader classification tasks. We conclude that pairwise discernment is a powerful tool for addressing data set imbalances and capturing meaningful patterns by reducing task complexity and improving model generalization.

\subsection{Sources of Error}
The models performed very well on majority classes but struggled significantly with minority classes. In this section, we discuss the potential sources of error contributing to this discrepancy and analyze where our approach fell short.

\subsubsection*{Imbalanced Data Set}
An early source of error in the project was the heavy class imbalance in the dataset. Nearly half of the training data belonged to one of the eight classes (see Table~\ref{tab:expression_classes}) (Happy), and  80\% of the data fell into the two majority classes (Happiness and Neutral). Consequently, the three minority classes combined to account for only  5\% of the training data. This imbalance occurred naturally due to the proportion of photos published "in the wild"; people were more likely to post photos of themselves when they were happy than when they were, for example, afraid.

\subsubsection*{Subjectivity of Facial Expressions}
The annotations provided by AffectNet were invaluable for training the model to recognize underlying patterns and quantify various facial expressions. However, a key challenge lies in the subjectivity of these annotations. According to the AffectNet paper: "To measure the agreement between the annotators, 36,000 images were annotated by two annotators. The annotations were performed fully blind and independently, i.e., the annotators were unaware of the intended query or the other annotator's response. The results showed that the annotators agreed on 60.7\% of the images." \cite{affectnet} This relatively low agreement rate highlights inconsistencies in how annotations were assigned, suggesting that the ground truth labels may not adhere to a consistent standard. Consequently, these inconsistencies could have disrupted the model's ability to reliably identify patterns in the data.

\subsection{Short Term Facial Emotion Recognition}

In the short term, Facial Emotion Recognition (FER) is expected to continue advancing significantly, driven by larger datasets with standardized annotations. While this project demonstrated relative success with a refined and intentional dataset, its performance "in the wild" remains unclear. Datasets like the CFP-FP dataset \cite{CFP}, which is used for facial verification from both frontal and profile angles, could be adapted for facial emotion recognition. Similarly, video datasets, such as the facial verification dataset Labeled Faces in the Wild (LFW) \cite{LFWTech}, could be developed for FER, enabling models to predict emotions dynamically. Real-time emotion detection has the potential to bring significant benefits, such as enhancing behavioral therapy by allowing researchers to process patient emotions quickly and autonomously.

Additionally, datasets and experiments based on annotations like Valence and Arousal, rather than rigid classes such as emotion expressions (e.g., happy, sad), would better capture the range of human emotions. Expression classes, as used in this project, remain too broad; for instance, an embarrassed laugh and a gleeful laugh are both categorized as "happy," yet they represent significantly different mental states.

\subsection{Long-Term Future for Machine Learning Emotion}
Image classification represents only the first step in enabling computers to understand and communicate emotion more broadly. The disparity between observable behavior and internal mental states remains substantial, and teaching models to classify behavior does not equate to teaching them emotion. Bayesian Neural Networks, with their capacity for world modeling, offer promising progress toward representing and communicating more complex mental and emotional states.

% or


\appendices

\ifCLASSOPTIONcompsoc
  \section*{Acknowledgments}
  We are grateful to Dr. Mohammad Mehdi Hosseini and the University of Denver Computer Science department for granting us access to the AffectNet database. 
  
  We appreciate the technical assistance provided by Amy Bush and the UTCS help staff, who gave us access to the compute and resources necessary for this paper.

\ifCLASSOPTIONcaptionsoff
  \newpage
\fi



%\bibliographystyle{IEEEtran}
\bibliographystyle{IEEETran}
\bibliography{references}

\begin{IEEEbiographynophoto}{Dylan Waldner}
Dylan Waldner received his B.A. in Philosophy from the University of Texas at Austin in 2024. He focused his research on the ethical design of intelligent agents, arguing that ethics underpin AI alignment solutions. Currently, he is completing research in neuroevolutionary games where a Bayesian Neural Network navigates a text-based choose-your-own-adventure game. His research interests include affective computing applications in ethical agent design. 
\end{IEEEbiographynophoto}
\begin{IEEEbiography}[{\includegraphics[width=1in,height=1.25in,clip,keepaspectratio]{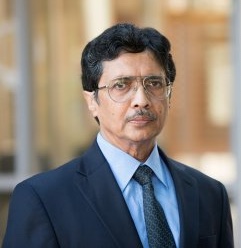}}]{Shyamal Mitra}
Shyamal Mitra received his Ph.D. in Astronomy from the University of 
Texas at Austin in 1988. He did postdoctoral work in the Binary Black
Hole Grand Challenge Project. He was a Research Associate at the Texas
Advanced Computing Center. He is currently an Associate Professor of
Instruction in the Computer Science department at the University of
Texas at Austin. His research interests include the application of machine
learning to scientific computations.
\end{IEEEbiography}


%\newpage

%\vfill


\end{document}